# Voxel-wise Weighted MR Image Enhancement using an Extended Neighborhood Filter

Joseph Suresh Paul, Joshin John Mathew, Souparnika Kandoth Naroth, and Chandrasekar Kesavadas

## Abstract

We present an edge preserving and denoising filter for enhancing the features in images, which contain an ROI having a narrow spatial extent. Typical examples include angiograms, or ROI's spatially distributed in multiple locations and contained within an outlying region, such as in multiple-sclerosis. The filtering involves determination of multiplicative weights in the spatial domain using an extended set of neighborhood directions. Equivalently, the filtering operation may be interpreted as a combination of directional filters in the frequency domain, with selective weighting for spatial frequencies contained within each direction. The advantages of the proposed filter in comparison to specialized non-linear filters, which operate on diffusion principle, are illustrated using numerical phantom data. The performance evaluation is carried out on simulated images from BrainWeb database for multiple-sclerosis, acute ischemic stroke using clinically acquired FLAIR images and MR angiograms.

## 1. Introduction

Distortion in medical images occurs due to low resolution, higher levels of noise, low contrast, geometric de-formations and presence of imaging artifacts. These imperfections can be present in all imaging modalities including CT, Mammograms, Ultra sound and MR. In particular, CT, and mammograms exhibit low contrast for soft-tissues, and ultra sound produces very noisy images. In MR images, the contrast between structures is limited by parameters involved in the image acquisition process, and the physical properties representative of the tissue characteristics. For example, the contrast between two tissue regions reduces if their parameters lie in very close ranges. In addition, subtle variability of the individual tissue parameters gives rise to a form of internal noise, leading to a variance of intensities within each region. This further complicates the detection of boundaries that delineate the two regions. A latter case of interest occurs when the contrast between the two tissue regions is high for regions farther away from the boundary. In low-resolution images, the low contrast for the regions close to the boundary results in blurred edges between the two regions owing to partial-volume effects. The signal model representing partial-volume effect is given by [1]

$$S(x) = (O(x) + n_{bio}(x)) * h(x) + n_{MR}(x) \qquad (1)$$

where $S(x)$ is the measured signal, $O(x)$ is the idealized signal, $n_{bio}(x)$ is the internal noise term, $h(x)$ models the partial-volume effect, and $n_{MR}(x)$ represents the statistical noise. Each of the cases discussed above leads to two different classes of image processing problems. Whereas the first requires enhancement of the tissue contrast between the two regions in the presence of internal and machine generated noise, the second is equivalent to a problem relating to sharpening of the tissue boundaries in the presence of partial volume effects and noise sources.

Under noise free conditions, both the contrast enhancement between tissue regions as in case-1, and sharpening of edges in case-2 can be accomplished using gradient based methods or its variations [2-5]. These methods, however, show limited performance in the presence of partial volume effects and other noise sources.

Application of linear gradient filters for delineation of low contrast regions (case-1) in the presence of noise is illustrated using a synthetic image shown in Fig. 1(a). The regions correspond to those enclosed by outer and inner circular boundaries. With an added 4% noise, the difference between mean intensities of the two homogenous regions is kept at $5 \times 10^{-3}$. As shown in Fig.1(b), the linear gradient operator is able to detect the edge of outer circle, but fails to detect the edge of inner circle.

Fig. 1 Here

Delineation of boundaries in this situation is generally accomplished using specialized non-linear filters [6-8], which operate on the diffusion principle. Denoising is achieved in an iterative manner, whereby finer image in each scale space level is derived using the optical flow. The optical flow in turn is defined as the product of a scalar diffusivity function, and the local spatial gradient. The diffusivity is obtained by using different forms of weighting functions on the magnitude of the squared gradients, scaled by a smoothing constant ($\kappa$) [8]. Larger values of diffusivity resulting from smaller values of gradients are indicative of larger extent of smoothing in the intra-tissue regions. Higher values of gradients near the edges, therefore, result in lower extent of smoothing and preservation of structural information. Edge preservation becomes optimal when the optical flow is maximized. The gradient at which the optical flow becomes a maximum is in turn dependent on the value chosen for $\kappa$. Thus the filtering operation requires an appropriate choice for $\kappa$. Choice of $\kappa$ below an optimal range results in spurious edge formation. On the other hand, if the chosen value is too large, structural information will be lost due to extreme effects of smoothing. One of the main drawbacks of diffusion based filtering, therefore, consists of choosing an optimal $\kappa$. A second drawback results from the difficulty in edge preservation, when the noise level near the edges exceeds a threshold limit. This can be explained from the narrowing of the useful $\kappa$-range with increase in noise level. The effects of $\kappa$ range and noise level on the filtering operation are illustrated in Fig. 2.

Fig. 2 Here

Panels (a)-(c) represent three distinct κ ranges, and numberings 1-4 correspond to noise levels of 4-7%. The left panels (a1)-(a4) correspond to κ values below the optimal range, showing spurious edge formation with increasing noise levels. The middle and right panels represent the ideal and supra-threshold ranges. It is seen that when the noise level exceeds about 6%, the ideal κ range becomes too narrow, and begins to exhibit spurious edge formation. Though diffusion based filter achieves edge preservation and denoising, it fails to produce sufficient separation between mean intensities of the two tissue regions. This is important in situations when the spatial extent of one of the regions is narrow and streaked such as in angiograms, or spatially distributed in multiple locations and contained within the other region, such as in multiple-sclerosis.

For such applications, it is essential to perform spatial operations using an extended set of neighborhood directions as compared to the Von-Neumann and Moore neighborhoods [9-10]. In the proposed method, the extended neighborhood directions are selected by connecting the reference pixel to the first pixel along all radial directions inside a square lattice of size $w \times w$. This is illustrated in Fig. 3 for a square lattice of size $w$=5. The radial directions along with the corresponding neighborhood pixels are shown for all the four quadrants within the lattice.

Fig. 3 Here

For a given lattice, the filtering operation involves binary map generation along all the directions. The binary maps are generated using a threshold based comparison of the central pixel intensity within a local window representing the lattice, and its neighborhood intensities along all the directions. This is achieved by comparing the input image with images shifted along each direction by pre and post multiplying the input image with exponentiated versions of lower and upper shift matrices. The choice of lower or upper shift matrices as pre and post factors is determined by the quadrant associated with the radial direction within the lattice.

## 2. Method

### 2.1 Generation of Binary Maps

Application of shift operation on the input image ($I$) requires the dimension $N$ to be equal in both the row and column directions. This is achieved by padding zeroes along the dimension of lower size. The shift operation involves pre and post multiplication of $I$ with powers $l$ and $m$ of upper ($U$), and lower ($L$) shift-matrices of size N × N. The extended neighborhood directions are obtained by connecting each pixel to the first pixel along all radial directions inside a square lattice of size $w \times w$ centered on the pixel. The choice of upper, or lower shift matrices for pre and post multiplication will be different for each quadrant of the lattice. This is described in Table-1.

Table-1 Here

Shift operations for center-to-right, center-to-left, center-to-top and center-to-bottom, are expressed as $IL$, $IU$, $LI$ and $UI$ respectively. The neighborhood pixels in a $3 \times 3$ lattice is shown in Fig. 4.

Fig. 4 Here

The shift operation required to map each of the neighborhood pixel on to the image coordinates of the reference position (shaded dark) is shown alongside the respective neighbors. As an illustrative example, the shift operations using a 3×3 lattice on an image matrix $A$ of size 5×5 is described in Table-2.

$$A = \begin{bmatrix} 1 & 2 & 3 & 3 & 5 \\ 6 & 7 & 8 & 9 & 10 \\ 11 & 12 & 13 & 14 & 15 \\ 16 & 17 & 18 & 19 & 20 \\ 21 & 22 & 23 & 24 & 25 \end{bmatrix} \quad L = \begin{bmatrix} 0 & 0 & 0 & 0 & 0 \\ 1 & 0 & 0 & 0 & 0 \\ 0 & 1 & 0 & 0 & 0 \\ 0 & 0 & 1 & 0 & 0 \\ 0 & 0 & 0 & 1 & 0 \end{bmatrix} \quad U = \begin{bmatrix} 0 & 1 & 0 & 0 & 0 \\ 0 & 0 & 1 & 0 & 0 \\ 0 & 0 & 0 & 1 & 0 \\ 0 & 0 & 0 & 0 & 1 \\ 0 & 0 & 0 & 0 & 0 \end{bmatrix}$$

Table-2 Here

For each radial direction k, the neighborhood pixel mapped on to the reference position in $I$ is given by

$$J_k = \begin{cases} L^{l_k} IL^{m_k}, & \text{for radial direction 'k' in quadrant} - I \\ L^{l_k} IU^{m_k}, & \text{for radial direction 'k' in quadrant} - II \\ U^{l_k} IU^{m_k}, & \text{for radial direction 'k' in quadrant} - III \\ U^{l_k} IL^{m_k}, & \text{for radial direction 'k' in quadrant} - IV \end{cases} \quad (2)$$

The binary map $(B_k)$ along $k^{th}$ direction is obtained using a threshold based comparison of intensities in the original image $I$, and the neighborhood map along $k^{th}$ direction $J_k$ as

$$B_k(i,j) = \begin{cases} 1 & \text{if } (I(i,j) - J_k(i,j)) > \eta \\ 0 & \text{otherwise.} \end{cases} \quad (3)$$

where $\eta$ represents the threshold used for binary map generation.

## 2.2 Computation of Shift-exponents

As shown in Eq. (2), the exponents $l_k$, and $m_k$ depend on the radial direction $k$, and the lattice size $w$. Fig. 5(a) shows a lattice with the reference pixel located at (0, 0) in the local coordinate system. For any given radial direction originating from the reference pixel, the immediate neighbor is identified as the first pixel located along that direction. The exponents of the pre and post shift matrices for mapping the image coordinates of the immediate neighbor to that of the reference pixel is obtained using a neighborhood mask $E$ given by

$$E(l,m) = \begin{cases} 1 & \text{if } \gcd(l,m) = 1 \\ 0 & \text{otherwise,} \end{cases} \quad (4)$$

for $1 \le l, m \le n$, where $n=(w-1)/2$. Positions in $E$ having ones are considered to be immediate neighbors. Generation of $E$ for a lattice of size $5 \times 5$ is illustrated in Fig.5 (b)-(c).

Fig. 5 Here

This excludes the immediate neighbors along x and y-directions with reference to the reference pixel. The shift exponents for these positions remain independent of $w$ and do not require any special computational procedure. Therefore, the steps in the current approach include exponent computation for the remaining directions only. The procedure for extension of this method to the remaining quadrants of the lattice is illustrated in Fig. 6.

Fig. 6 Here

Sample neighborhood masks for various lattice sizes are shown in Table- 3.

Table-3 Here

The number of directions ($N_q$) in the first quadrant of the lattice, is therefore obtained by summing the unit elements of the neighborhood mask,

$$N_q = \sum_{l=1}^{n} \sum_{m=1}^{n} E(l,m)$$

(5)

Considering all four quadrants, the number of immediate neighbors of any pixel will, therefore, be $N_d=4\times(N_q+1)$. Once the shift exponents are computed, the $N_d$ numbers of binary maps can be generated as described in Eq. (3).

## 2.3 **Filtering Procedure**

A block schematic of the filtering procedure is shown in Fig. 7. Binary maps are generated for all directions, and threshold $\eta$ for intensity comparison of the input and shifted images is estimated using a threshold estimation algorithm, explained in a later section.

Fig. 7 Here

The weight for each pixel of the input image is obtained as the intensity at the corresponding position in the summation of binary maps (*BWI*) along all directions

$$BWI = \sum_{k=1}^{N_d} B_k$$

(6)

The filtered image is then estimated as

$$O(i, j) = I(i, j) + \big(I(i, j) \times BWI(i, j)\big)$$

(7)

## 2.4 Threshold Estimation

The threshold is calculated by first estimating the noise variance ($\sigma^2_M$) for a given Region-Of-Interest (ROI), using skewness ($\gamma$) of the intensity distribution of pixels within the ROI. The steps used for estimation of $\sigma^2_M$ is borrowed from [11], and summarized in the flowchart shown in Fig. 8.

Fig.8 Here

The threshold ($\eta$) is then chosen from a range of values $\eta \in [\sigma^2_M, C_{ROI}]$, where $C_{ROI}$ denotes difference between the mean intensities of the two tissue regions within the ROI. The effect of choosing a specific value in this range, on the filtered image is explained in section 3.2.

## 3. Result

## 3.1 Numerical Results

Fig. 9(a) shows a phantom image with two circular edges. The inner edge is not visible due to the mean intensity difference of 5 $\times 10^{-3}$. This image is filtered using different lattice sizes of *w*=3, 7, 11, and 15. For small values of *w,* the filter operates as an edge detector. As *w* increases, the spread of the edge extends to the centroid of the region having a higher mean intensity. This effect is illustrated in Figs. 9(g)-(j). The corresponding filtered versions are shown in panels (b)-(e). It is seen that as $w$ is increased beyond 11, the edges spread from opposite directions, so as to fill the entire region with a mean intensity larger than its surrounding pixels. In the context of medical images, these regions can be likened to those of small sized lesions, such as multiple sclerosis seen in PD, or T1-weighted MR images. This is true, especially, for modalities in which the lesions exhibit a slightly larger intensity than the surrounding healthy region. As the lesion size increases, the lattice size required to enhance, or fill the lesion, will accordingly be higher. Increasing the lattice size is also accompanied by an increase in the $C_{ROI}$, equivalent to the difference between the mean intensities of the two regions as evident from panels (i) and (j). This example illustrates an implementation of case-1, as discussed section-1.

Fig.9 Here

## 3.2 Effect of $\eta$

As discussed in section 2.4, the threshold ($\eta$) is selected in such a way that it should be less than the difference between the means of two tissue regions in the ROI ($C_{ROI}$), and greater than $\sigma^2_M$. The effect of choosing an appropriate $\eta$ is illustrated using the phantom image in Fig. 10, and a sample MR image with MS lesions obtained from the BrainWeb database [12]. The ROI in Fig. 10 is chosen to be the circular region, encompassing an inner disk of a slightly higher mean intensity, as explained earlier in sections-1 and 3.1. For a 1% added noise, the $\sigma^2_M$ in the ROI is estimated to be $4.2903\times10^{-5}$, using the procedure outlined in section 2.4. The panels (b)-(d) represent filtered images corresponding to $\eta < \sigma^2_M$, $\sigma^2_M < \eta < C_{ROI}$, and $\eta > C_{ROI}$ respectively.

Fig.10 Here

The condition $\eta < \sigma^2{}_M$ results in spurious edges as shown in Fig. 10(b). Likewise, for $\eta > C_{ROI}$, the contrast between two regions is not enhanced as shown in Fig. 10(d). This is easily deduced from the fact that the weights obtained from binary maps will be close to zero. The optimal performance is shown in Fig. 10(c). The ROI for the BrainWeb image is the bounding box shown in red in Fig. 11(a). Filtering is performed on PD images of slice thickness 1mm, intensity non-uniformity of 20% and different levels of added noise 1-7% in steps of 2%. The filtered images corresponding to 1% noise level are shown in Figs. (b)-(d) for $\eta < \sigma^2{}_M$, $\sigma^2{}_M < \eta < C_{ROI}$, and $\eta > C_{ROI}$ respectively. The insets (f-h) show the filtered versions of the ROI. The results obtained are identical to the simulated example in Fig. 10.

Fig.11 Here

### 3.3 <u>Application to Simulated Images</u>

In this section, the effect of applying binary weighted maps on simulated PD weighted images with MS lesions is illustrated. Sample PD images of slice thickness 1mm, intensity non-uniformity of 20% and different levels of added noise (1-7 % in steps of 2%) are taken from BrainWeb database. The lattice size used for filtering is chosen based on the maximum size of MS lesions. The threshold is estimated using the estimation procedure outlined in section 2.4. In the input image shown in Fig. 12(a), the MS lesions are not clearly visible due to low contrast from the surrounding tissue. The filtered image is shown in Fig. 12(d). The panels (b) and (e) show regions within the ROI highlighted by the bounding boxes in the input and filtered images respectively. The plots of intensity variation across the lesion in the original and filtered images are shown in the adjoining panels (c) and (f).

Fig.12 Here

The effect of increasing the lattice size is shown in Fig. 13. It is seen that for the given ROI of the BrainWeb sample image, an acceptable performance is achieved for a lattice of 11. With further increase in lattice size, there is no marked difference in performance.

Fig.13 Here

## 3.4 <u>Application to Clinical images</u>

Fig. 14 illustrates the filtering applied to a sample FLAIR image acquired on 1.5T clinical MR scanner (Magnetom- Avanto, Siemens, Erlangen, Germany) with a 12 channel head coil. The MRI parameters for FLAIR sequence includes the TE : 108 milliseconds; TR : 8140 milliseconds; TI : 2500 milliseconds; field of view : 230mm; matrix 256 ×184; 24 slices; 5mm slice thickness, 30% gap). The patient's MRI FLAIR image shows an infarct in the left insular cortex and frontal operculum. The ROI indicating the area of stroke is highlighted using a bounding box in this image. The regions within the bounding box of the original and filtered images are shown in panels (c) and (d) respectively. It is seen that the filtered image highlights the spatial profile of the thrombotic vasculature. The surrounding regions of edema are seen with darker shades in the filtered image. After filtration, brain shows better gray- white distinction. The area of infarct shows various intensities within, probably representing the extent of infarct severity. The margins of the infarct are better made out.

Fig.14 Here

Effect of lattice size is shown in Fig. 15. An optimal performance is achieved for $w \geq 15$.

Fig. 15 Here

## 3.5 <u>Performance Characterization</u>

The synthetic image described in section 3.1 is used for calculating the Contrast-to-Noise Ratio (CNR) for both the binary weighted filter, and anisotropic filters described by Perona and Malik [6], and Jiang et al., [8]. CNR between two regions a and b ($CNR_{a/b}$) is calculated as $(S_a$-$S_b)/\sigma$, where $S_a$ and $S_b$ refer to the average signal intensity of brighter, and darker regions respectively. The CNR is plotted against noise standard deviation ($\sigma$) in Fig. 16.

Fig.16 Here

For both the proposed and anisotropic filters, the CNR is observed to decrease with increase in noise variance. The performance of proposed method is seen to be better when the standard deviation of intensity variation is less than 0.03 corresponding to a mean intensity of 0.655. When the standard deviation exceeds 5% of the mean intensity, a pre-processing step using Non-local means (NLM) filter [13] is suggested.

Parameters for NLM filtering include the search window size ($t$), similarity window size ($f$) and a weight decay control parameter ($h$) [13]. The de-noising capability of NLM filter is controlled by the third parameter $h$, identical to the estimated noise variance $\sigma^2_M$. The steps for pre-filtering using NLM are shown in Fig. 17.

Fig.17 Here

The result of filtering a BrainWeb image sample with 5% added noise and 3mm slice thickness is shown in Fig. 18. The panels (a)-(b) correspond to the noisy input image with 5% added noise, image processed using NLM filter with search window size $t$= 5, similarity window size $f$ =1, and $h$=11. The image in Fig. 18(b) is further processed using the proposed filter with a lattice of 17. The resulting image is shown in panel (c). The zoomed versions of the ROI are shown in panels (d)-(f).

Fig.18 Here

The MR angiography image in Fig. 19 after filtration, shows an enhanced version of the peripheral vasculature.

Fig.19 Here

## 4. **Discussion**

The proposed filter is useful for enhancement of lesions or structures with a higher intensity with respect to its surroundings. The filter is derived from a sequence of binary maps estimated using an intensity threshold based comparison of the input image, and its spatially shifted versions. The shifting operation is performed along a series of extended neighborhood directions determined from a lattice of a predetermined size $w \times w$. The filtering involves determination of multiplicative weights in the spatial domain, derived from the binary maps. The whole procedure may be summarized in terms of deriving a frequency domain kernel, obtained using the DFT of the weight matrix. The kspace of the filtered image is computed by convolving the raw kspace with the weight matrix DFT. The kspaces of the filter weights corresponding to each of the extended directions for a lattice size of $3 \times 3$ is shown in Fig. 20.

Fig.20 Here

It is observed that each of the filtered kspaces are oriented along specific angular directions. Equivalently, the filtering operation may be interpreted as a directional filter with selective weighting for the spatial frequencies contained within the specific angular band for that direction.

# Figures

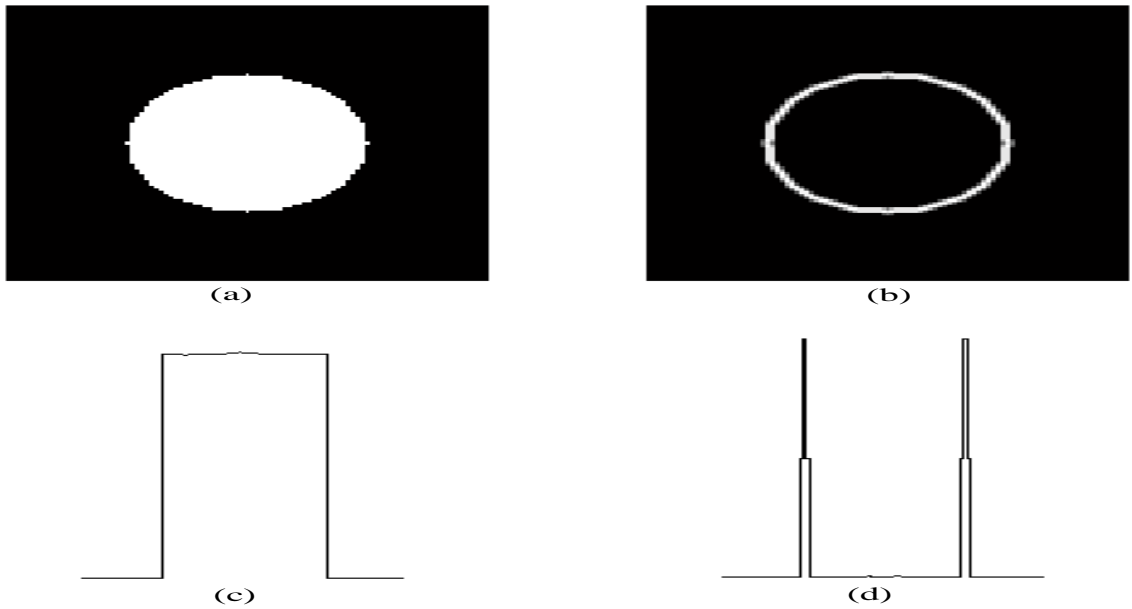

Fig. 1: (a) Synthetic image with 4% added noise, (b) edge filtered Image, (c)-(d) Cross-sections of (a) and (b).

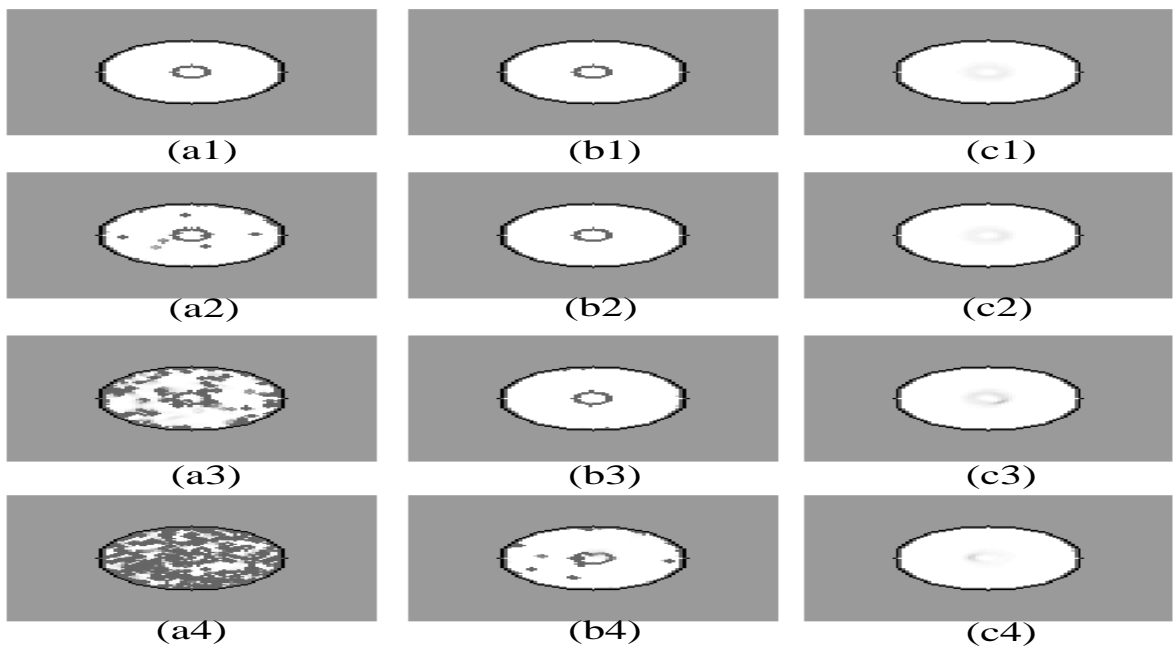

**Fig. 2: (a)-(c) Anisotropic filtering using different smoothing constants (κ) ,(a1)-(a4) Filtered images for noise levels 4-7% and κ below optimal range, (b1)-(b4) Filtered images for noise levels 4-7% and □ within optimal range, (c1)-(c4) Filtered images for noise levels 4-7% and κ above optimal range.**

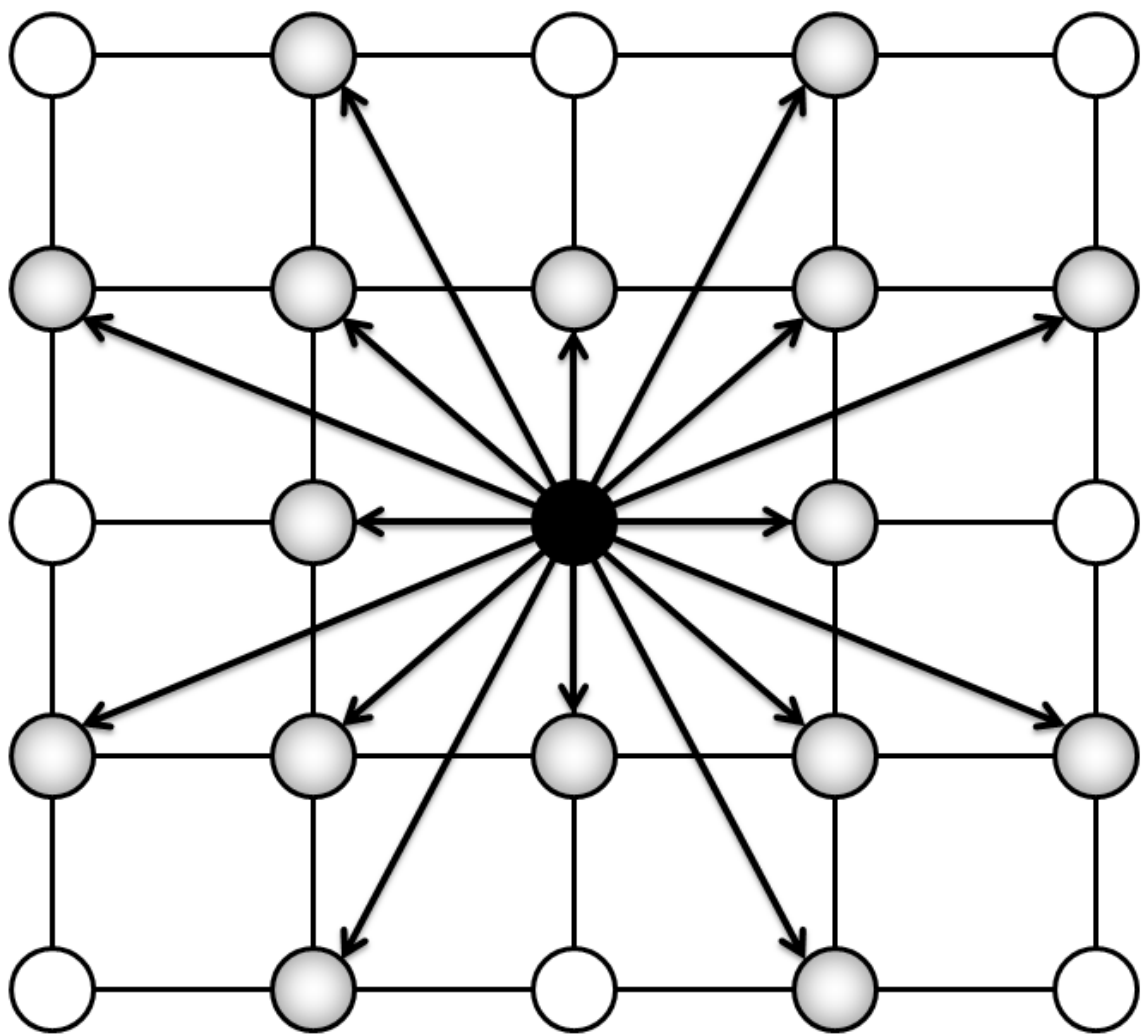

Fig. 3: Extended neighborhood pixels along different directions for a lattice of size 5×5.

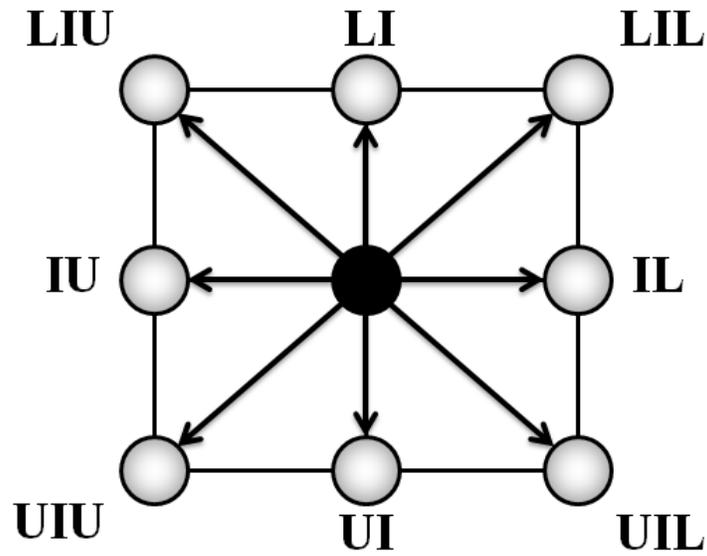

**Fig. 4: Shift operations for mapping neighborhood pixels for a lattice of size 3×3.**

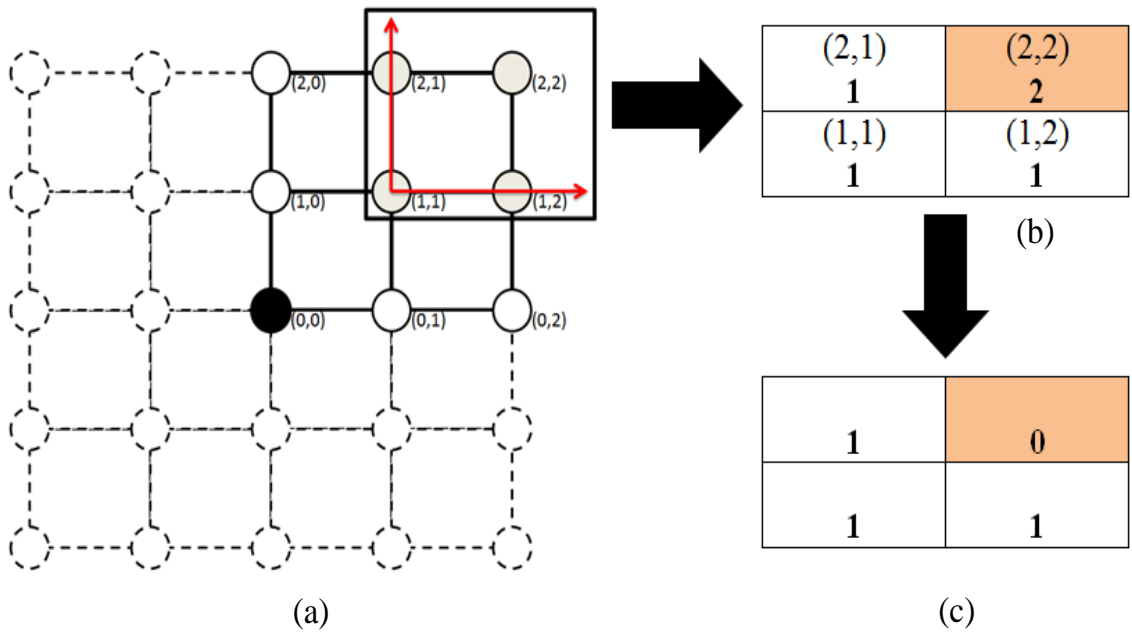

**Fig. 5: Generation of Neighborhood Mask E. (a) Bounding box showing pixels in the first quadrant of the square lattice, (b) Local co-ordinates of the first quadrant with corresponding GCD values, (c) Neighborhood mask E.**

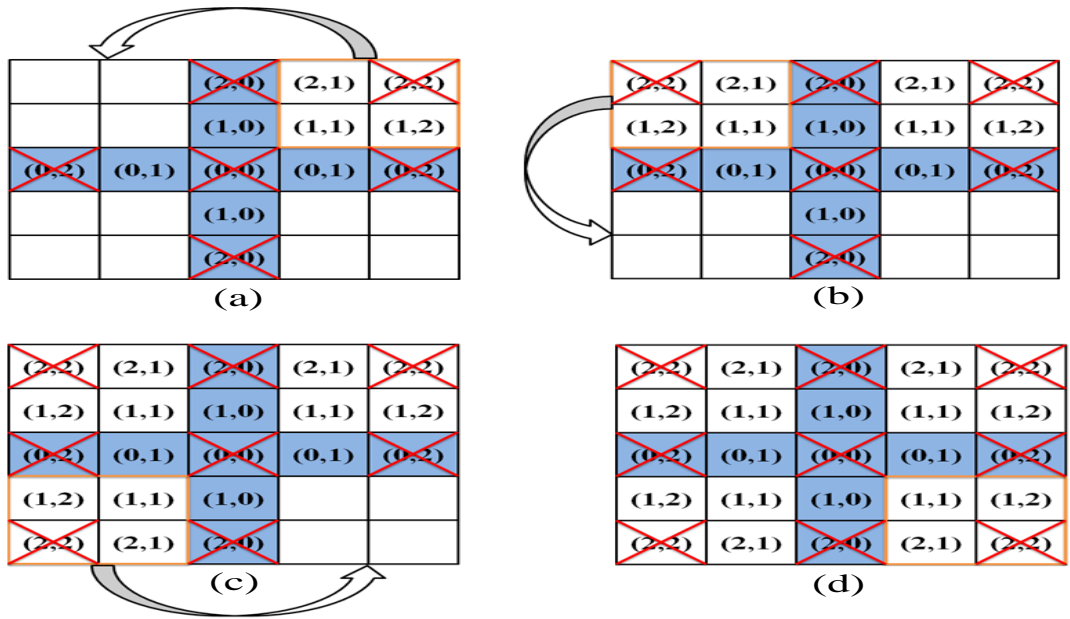

Fig. 6: Procedure for extension of exponent computation to the remaining quadrants of the lattice.

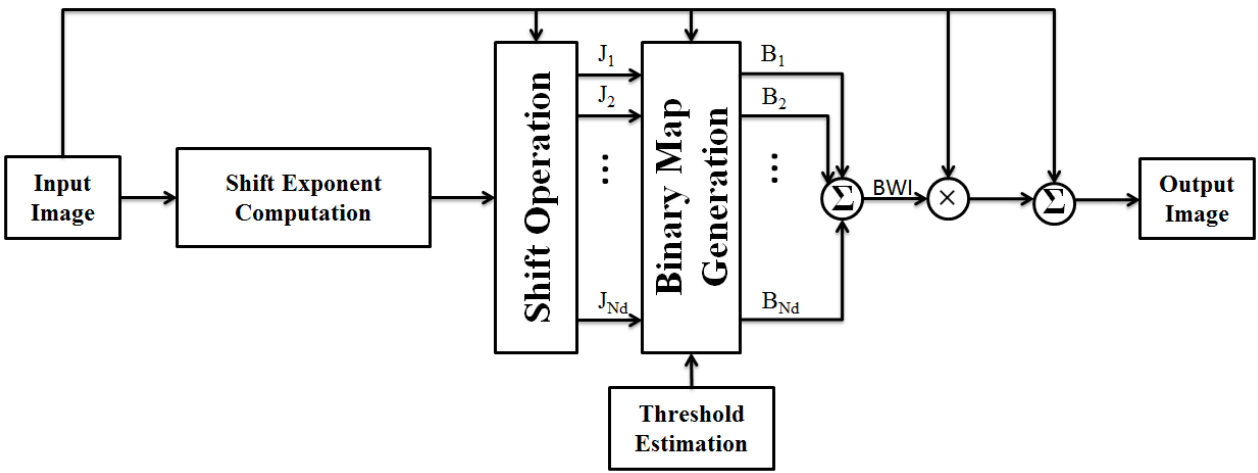

Fig. 7: Block Diagram of the Binary Weighted Filter.

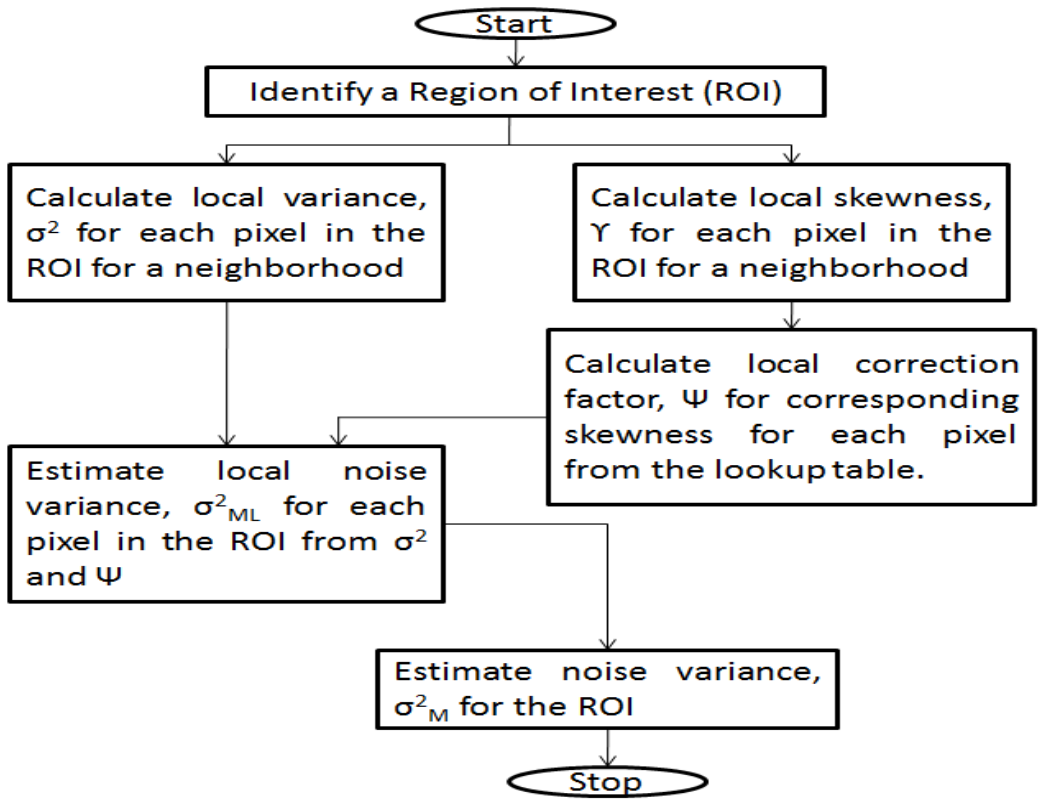

**Fig. 8: Estimation of Noise variance.**

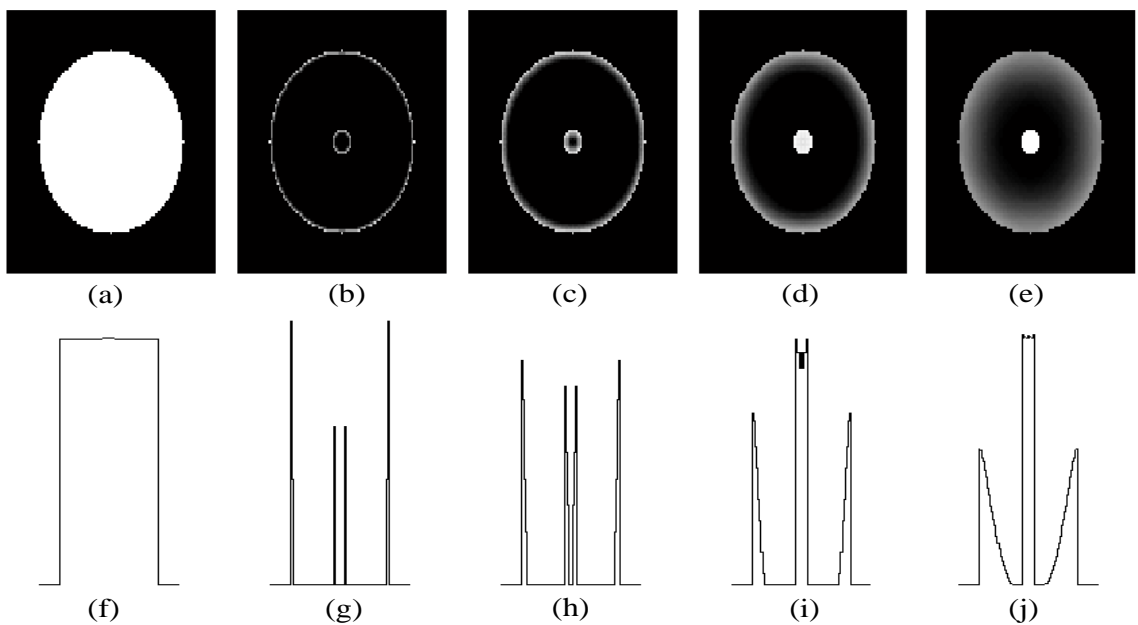

**Fig. 9: (a) Synthetic image, (b)-(e) Summation of binary maps for lattice size 3, 7, 11 and 15, (f)-(j) Cross-sections of (a)-(e).**

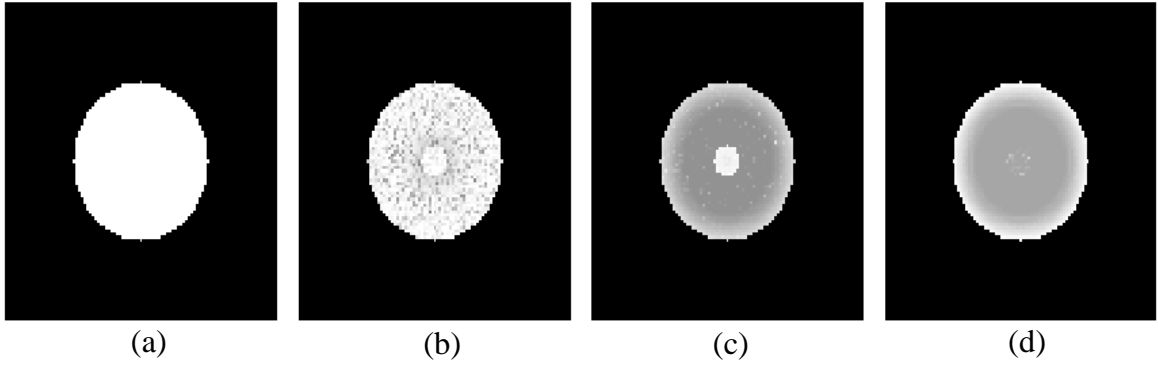

**Fig. 10: (a) Synthetic image consisting of two concentric discs, with mean intensity difference of annular regions represented by $C_{ROI}$ and variance $\sigma^2_M$, (b) Image filtered using threshold $\eta < \sigma^2_M$, (c) Image filtered using $\sigma^2_M < \eta < C_{ROI}$, (d) Image filtered using $C_{ROI} < \eta$.**

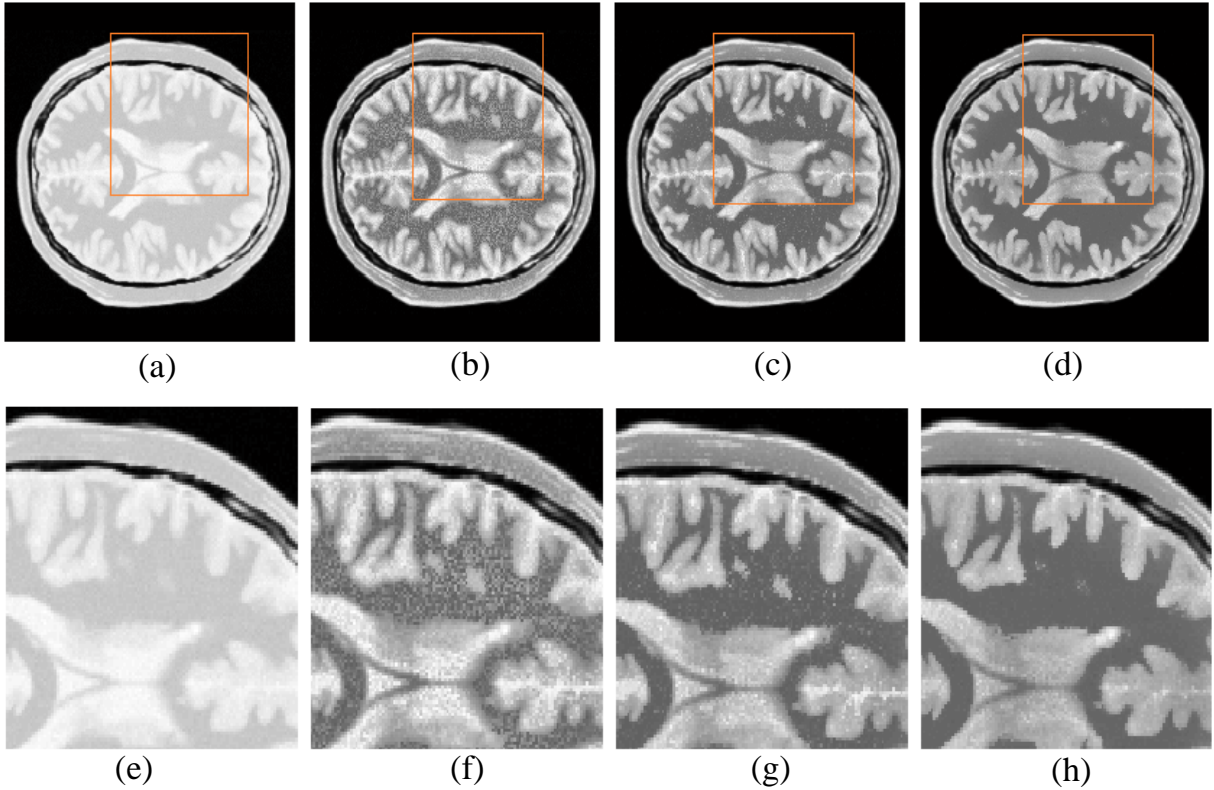

**Fig. 11: (a) A sample slice of simulated PD weighted image from BrainWeb database with added 1% noise level, (b) Image filtered using threshold less than $\sigma^2_M$, (c) Image filtered using threshold greater than $\sigma^2_M$ and less than $C_{ROI}$, (d) Image filtered using threshold greater than $C_{ROI}$, (e)-(h) Regions inside bounding boxes in (a)-(d).**

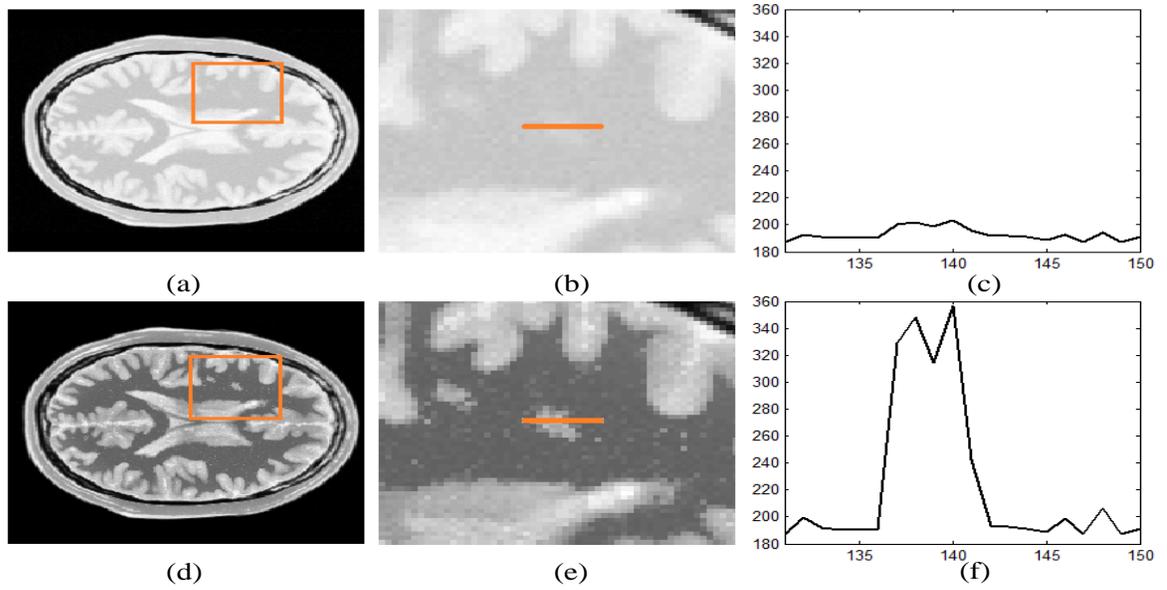

**Fig. 12:** (a) A sample slice of simulated PD weighted image from BrainWeb database with added 1% noise level, (d) Filtered image, (b) & (e) Region inside the bounding boxes in (a) & (d), (c) & (f) Intensity variation across the horizontal bars shown in (b) & (e).

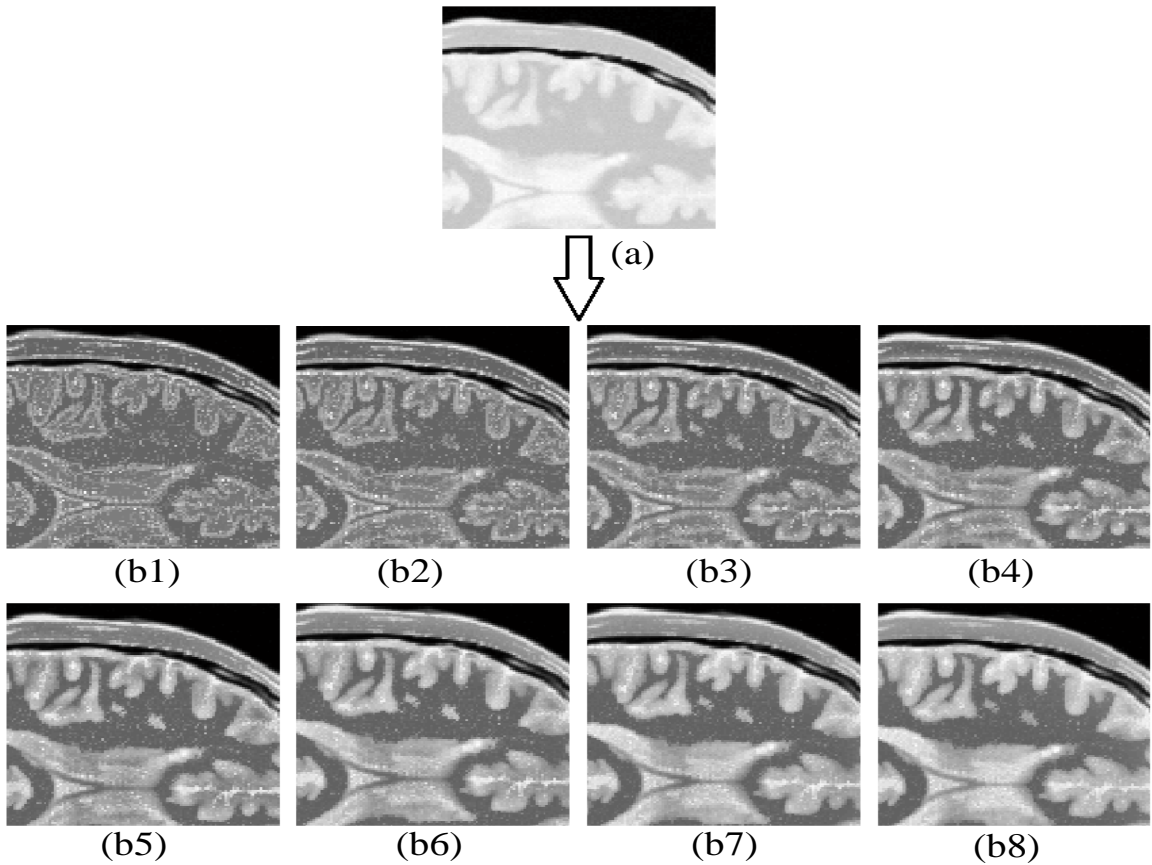

**Fig. 13:** Effect of Lattice size, (a) ROI of the input image with 1% noise level, (b1)-(b3) ROI in filtered images for lattice sizes 3, 5, 7, 9, 11, 13, 15, and 17 respectively.

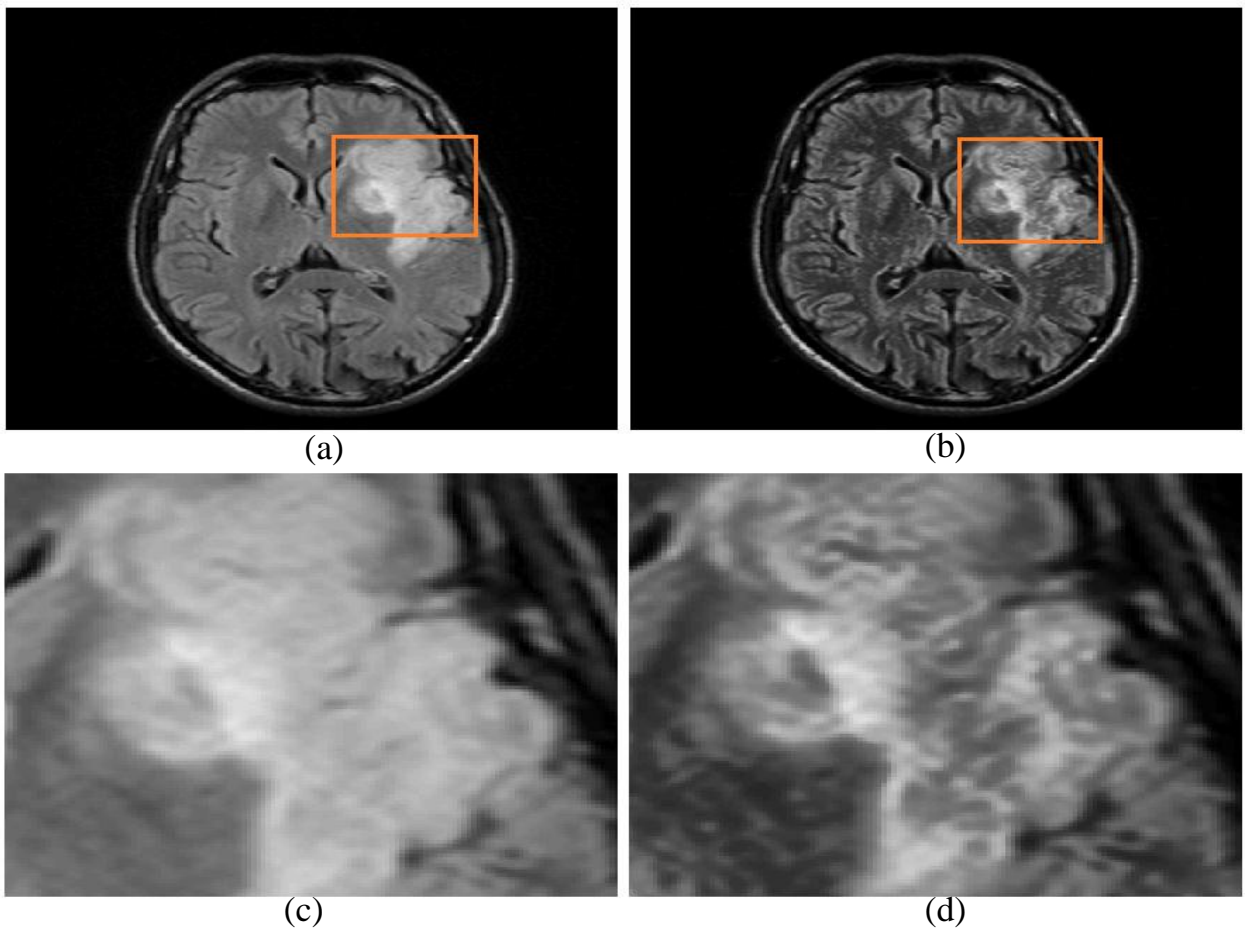

(a)  (b)

(c)  (d)

**Fig. 14: (a) FLAIR image showing an area of stroke, (b) Filtered Image, (c) Region inside bounding box in (a), (d) Region inside bounding box in (b).**

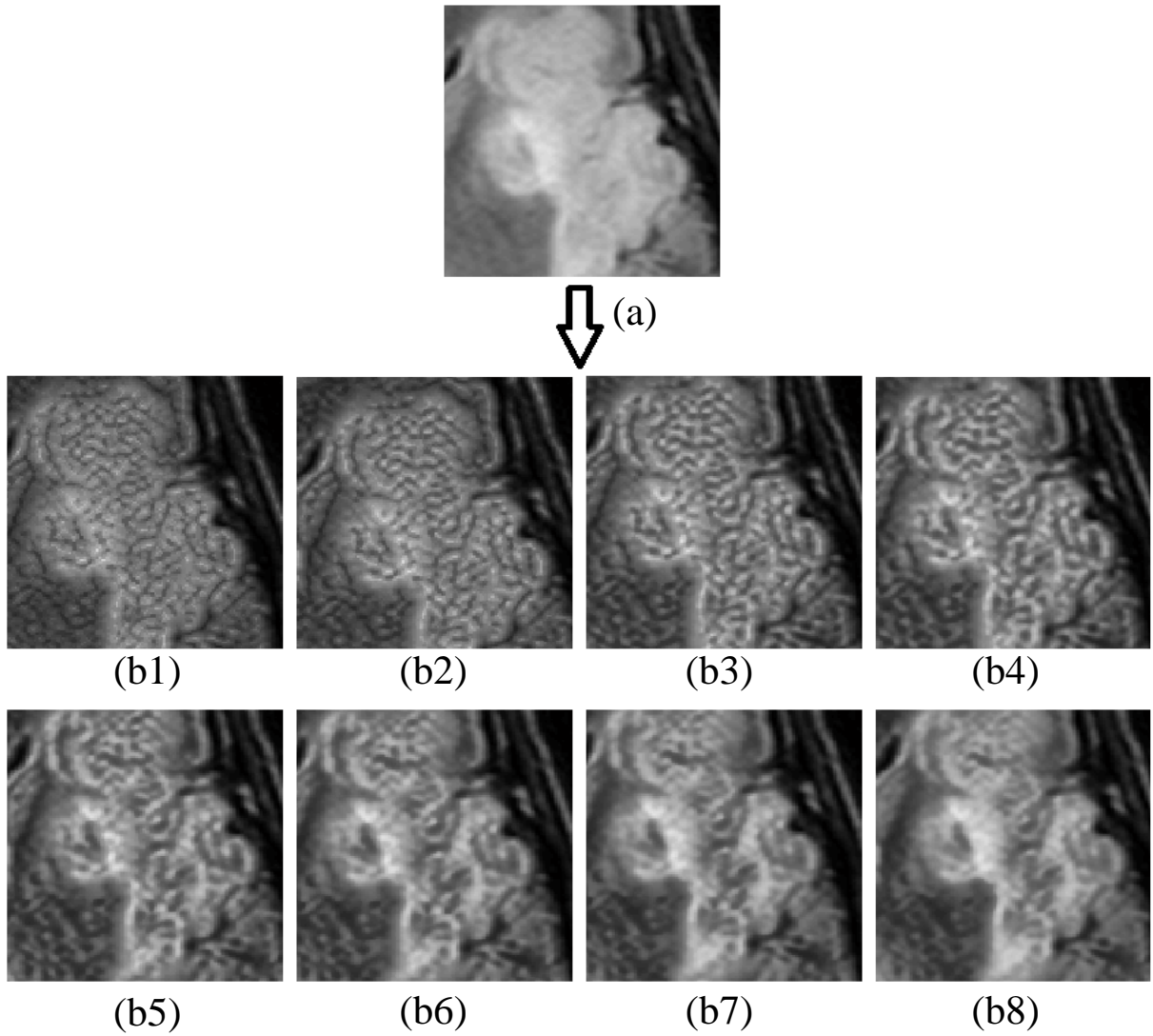

**Fig. 15: Effect of Lattice size, (a) ROI of the FLAIR image, (b1)-(b8) ROI in filtered images for lattice sizes 3, 5, 7, 9, 11, 13, 15, and 17 respectively.**

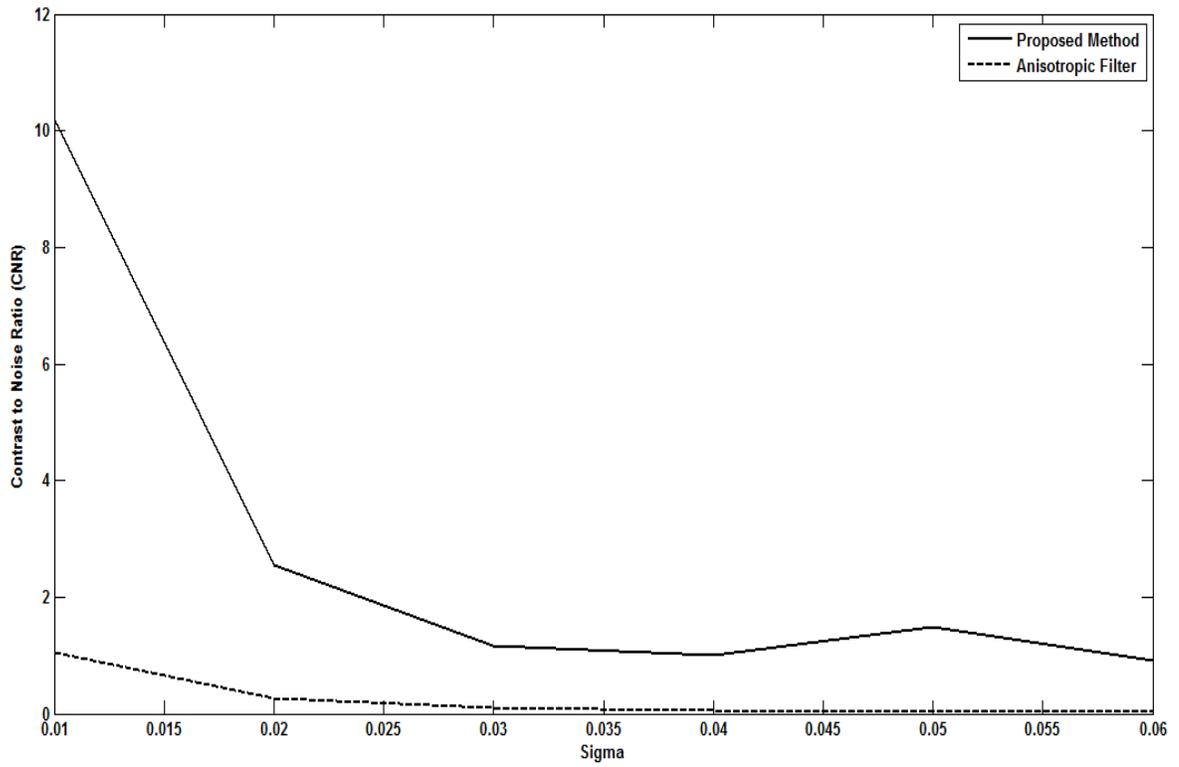

Fig. 16: Plots of CNR against noise standard deviation (σ) for the proposed versus Anisotropic filter.

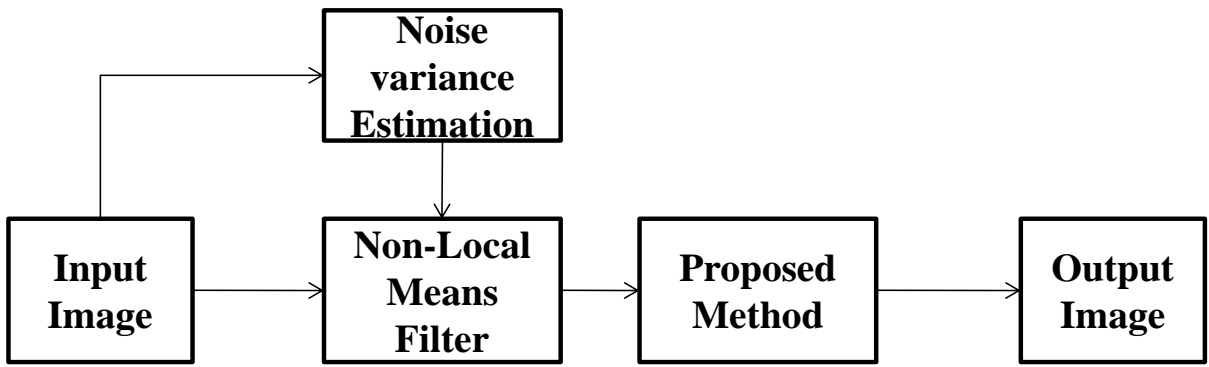

Fig. 17: Pre-filtering using Non-Local Means (NLM) filter.

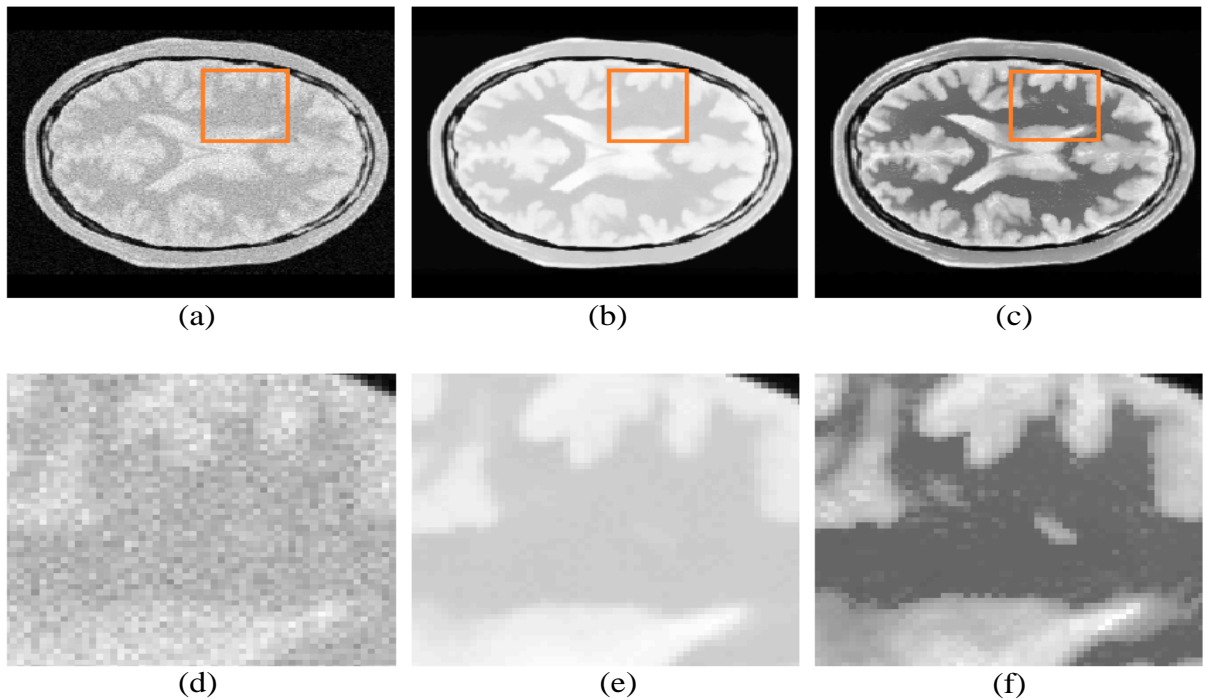

**Fig. 18:** (a) A sample slice of simulated PD weighted image from BrainWeb database with added 5% noise level, (b) NLM filtered image, (c) The result of proposed binary weighted filter applied to the NLM filtered input image, (d)-(f) Regions within the bounding boxes in (a)-(c).

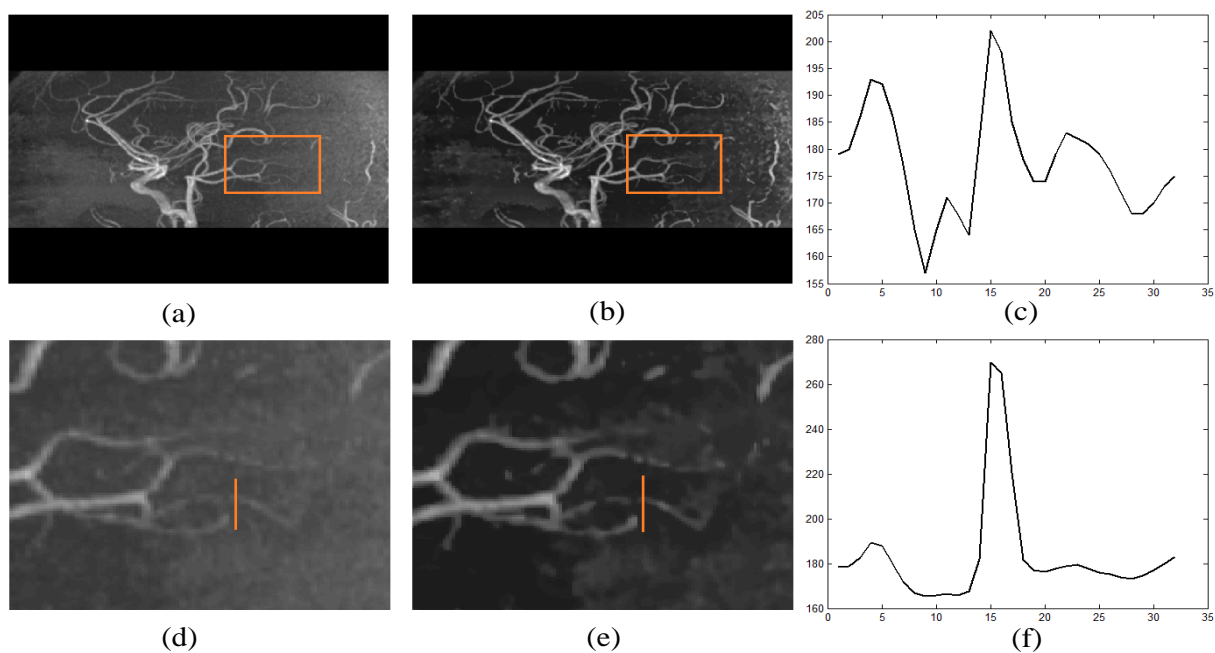

**Fig. 19:** (a) A sample slice of MR Angiogram, (b) Filtered image, (d)-(f): Regions within the bounding boxes in (a) & (b), (c) & (f): Intensity variation across the vertical bars shown in (d) & (e).

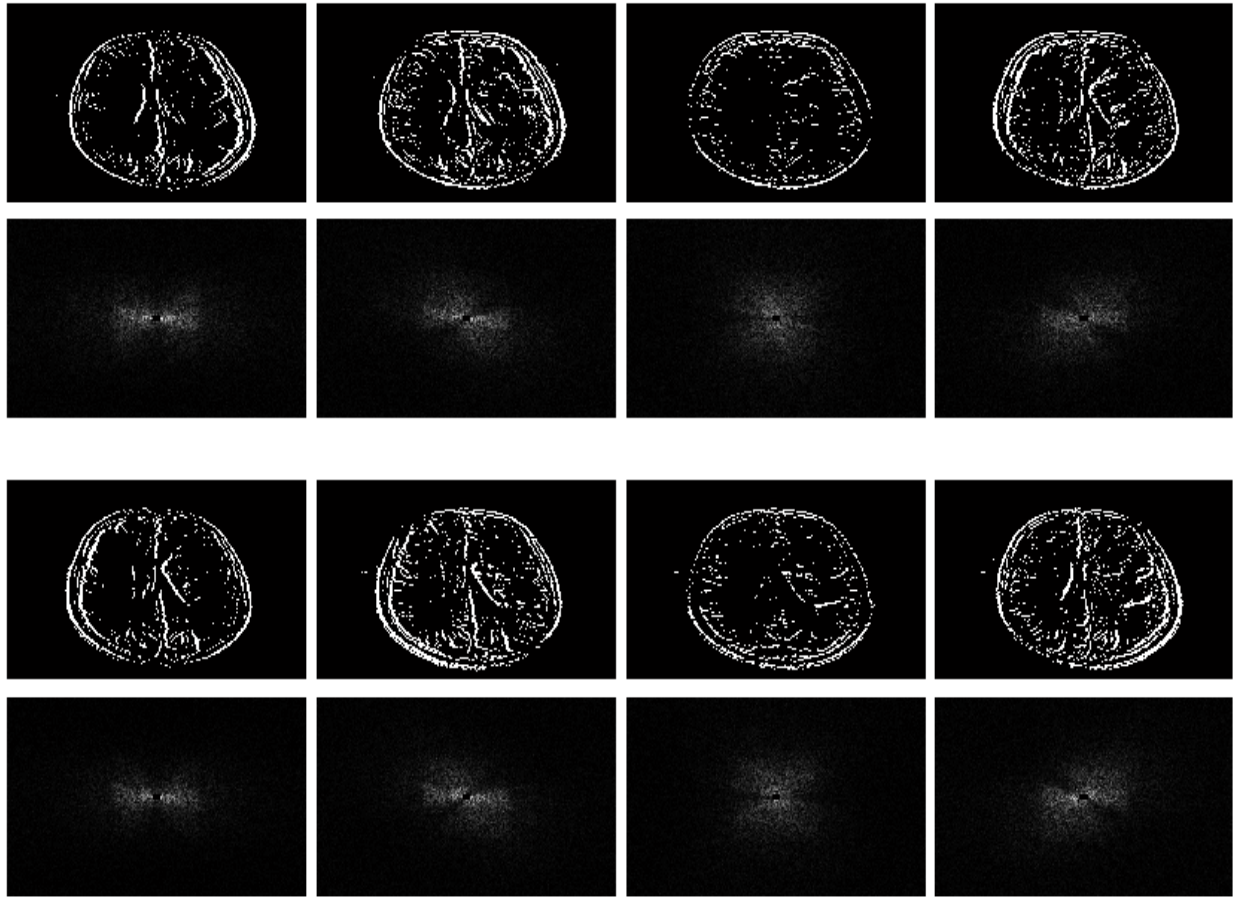

Fig. 20: Binary maps and corresponding frequency domain images for 8 directions.

# Tables

| Quadrant | Shift Operation |
|----------|-----------------|
| I$^{st}$ | $L^l \, I \, L^m$ |
| II$^{nd}$ | $L^l \, I \, U^m$ |
| III$^{rd}$ | $U^l \, I \, U^m$ |
| IV$^{th}$ | $U^l \, I \, L^m$ |

**Table 1: Pre and Post multiplication shift operators for each quadrant in the lattice.**

$$AL = \begin{bmatrix} 2 & 3 & 4 & 5 & 0 \\ 7 & 8 & 9 & 10 & 0 \\ 12 & 13 & 14 & 15 & 0 \\ 17 & 18 & 19 & 20 & 0 \\ 22 & 23 & 24 & 25 & 0 \end{bmatrix}$$

$$LAL = \begin{bmatrix} 0 & 0 & 0 & 0 & 0 \\ 2 & 3 & 4 & 5 & 0 \\ 7 & 8 & 9 & 10 & 0 \\ 12 & 13 & 14 & 15 & 0 \\ 17 & 18 & 19 & 20 & 0 \end{bmatrix}$$

$$LA = \begin{bmatrix} 0 & 0 & 0 & 0 & 0 \\ 1 & 2 & 3 & 4 & 5 \\ 6 & 7 & 8 & 9 & 10 \\ 11 & 12 & 13 & 14 & 15 \\ 16 & 17 & 18 & 19 & 20 \end{bmatrix}$$

$$LAU = \begin{bmatrix} 0 & 0 & 0 & 0 & 0 \\ 0 & 1 & 2 & 3 & 4 \\ 0 & 6 & 7 & 8 & 9 \\ 0 & 11 & 12 & 13 & 14 \\ 0 & 16 & 17 & 18 & 19 \end{bmatrix}$$

$$AU = \begin{bmatrix} 0 & 1 & 2 & 3 & 4 \\ 0 & 6 & 7 & 8 & 9 \\ 0 & 11 & 12 & 13 & 14 \\ 0 & 16 & 17 & 18 & 19 \\ 0 & 21 & 22 & 23 & 24 \end{bmatrix}$$

$$UAU = \begin{bmatrix} 0 & 6 & 7 & 8 & 9 \\ 0 & 11 & 12 & 13 & 14 \\ 0 & 16 & 17 & 18 & 19 \\ 0 & 21 & 22 & 23 & 24 \\ 0 & 0 & 0 & 0 & 0 \end{bmatrix}$$

$$UA = \begin{bmatrix} 6 & 7 & 8 & 9 & 10 \\ 11 & 12 & 13 & 14 & 15 \\ 16 & 17 & 18 & 19 & 20 \\ 21 & 22 & 23 & 24 & 25 \\ 0 & 0 & 0 & 0 & 0 \end{bmatrix}$$

$$UAL = \begin{bmatrix} 7 & 8 & 9 & 10 & 0 \\ 12 & 13 & 14 & 15 & 0 \\ 17 & 18 & 19 & 20 & 0 \\ 22 & 23 & 24 & 25 & 0 \\ 0 & 0 & 0 & 0 & 0 \end{bmatrix}$$

**Table 2: Matrix 'A' shifted along 8 directions.**

**Lattice Size $w \times w$ = 3×3**

gcd(l,m):

| (1,1) 1 |
|---|

Neighbourhood Mask E:

| 1 |
|---|

---

**Lattice Size 5×5**

gcd(l,m):

| (2,1) 1 | (2,2) 2 |
|---|---|
| (1,1) 1 | (1,2) 1 |

Neighbourhood Mask E:

| 1 | 0 |
|---|---|
| 1 | 1 |

---

**Lattice Size 7×7**

gcd(l,m):

| (3,1) 1 | (3,2) 1 | (3,3) 3 |
|---|---|---|
| (2,1) 1 | (2,2) 2 | (2,3) 1 |
| (1,1) 1 | (1,2) 1 | (1,3) 1 |

Neighbourhood Mask E:

| 1 | 1 | 0 |
|---|---|---|
| 1 | 0 | 1 |
| 1 | 1 | 1 |

---

**Lattice Size 9×9**

gcd(l,m):

| (4,1) 1 | (4,2) 2 | (4,3) 1 | (4,4) 4 |
|---|---|---|---|
| (3,1) 1 | (3,2) 1 | (3,3) 3 | (3,4) 1 |
| (2,1) 1 | (2,2) 2 | (2,3) 1 | (2,4) 2 |
| (1,1) 1 | (1,2) 1 | (1,3) 1 | (1,4) 1 |

Neighbourhood Mask E:

| 1 | 0 | 1 | 0 |
|---|---|---|---|
| 1 | 1 | 0 | 1 |
| 1 | 0 | 1 | 0 |
| 1 | 1 | 1 | 1 |

---

**Lattice Size 11×11**

gcd(l,m):

| (5,1) 1 | (5,2) 1 | (5,3) 1 | (5,4) 1 | (5,5) 5 |
|---|---|---|---|---|
| (4,1) 1 | (4,2) 2 | (4,3) 1 | (4,4) 4 | (4,5) 1 |
| (3,1) 1 | (3,2) 1 | (3,3) 3 | (3,4) 1 | (3,5) 1 |
| (2,1) 1 | (2,2) 2 | (2,3) 1 | (2,4) 2 | (2,5) 1 |
| (1,1) 1 | (1,2) 1 | (1,3) 1 | (1,4) 1 | (1,5) 1 |

Neighbourhood Mask E:

| 1 | 1 | 1 | 1 | 0 |
|---|---|---|---|---|
| 1 | 0 | 1 | 0 | 1 |
| 1 | 1 | 0 | 1 | 1 |
| 1 | 0 | 1 | 0 | 1 |
| 1 | 1 | 1 | 1 | 1 |

**Table 3: Neighborhood Mask generation for lattice sizes, 3, 5, 7, and 11.**